\documentclass[conference]{IEEEtran}

\usepackage[printwatermark]{xwatermark}
\usepackage{graphicx}
\usepackage{array}
\usepackage{float}
\usepackage{nicefrac}
\usepackage[table]{xcolor}
\usepackage{booktabs}
\usepackage{amssymb}
\usepackage{pifont}

\usepackage{hyperref}
\hypersetup{
	colorlinks   = true, 
	urlcolor     = blue, 
	linkcolor    = blue, 
	citecolor   = red 
}

\newcommand{\MYfooter}{\smash{
\hfil\parbox[t][\height][t]{\textwidth}{}\hfil\hbox{}}}
\makeatletter
\def\ps@IEEEtitlepagestyle{%
\def\@oddhead{\mbox{}December 2017\rightmark \hfil }%
\def\@oddfoot{\MYfooter}%
\def\@evenfoot{\MYfooter}}
\makeatother
\usepackage{multirow}

\usepackage[switch]{lineno}

\begin{document}
	
\title{DeepGestalt - Identifying Rare Genetic Syndromes Using Deep Learning}

\author{
	\IEEEauthorblockN {
		Yaron Gurovich$^1$, 
		Yair Hanani$^1$,
		Omri Bar$^1$,
		Nicole Fleischer$^1$,
		Dekel Gelbman$^1$,
		Lina Basel-Salmon$^{2,3}$, \\
		Peter Krawitz$^4$,
		Susanne B Kamphausen$^5$,
		Martin Zenker$^5$,
		Lynne M. Bird$^{6,7}$, 
		Karen W. Gripp$^8$
	}
	
	\IEEEauthorblockA { \\
		$^{1}$FDNA Inc., Boston, Massachusetts, USA \\
		$^{2}$Sackler Faculty of Medicine, Tel Aviv University, Tel Aviv, Israel \\
		$^{3}$Recanati Genetic Institute, Rabin Medical Center \& Schneider Children's Medical Center, Petah Tikva, Israel \\
		$^{4}$Institute for Genomic Statistic and Bioinformatics, University Hospital Bonn, \\ Rheinische-Friedrich-Wilhelms University, Bonn, Germany \\
		$^{5}$Institute of Human Genetics, University Hospital Magdeburg, Magdeburg, Germany    \\
		$^{6}$Department of Pediatrics, University of California, San Diego, California, USA \\
		$^{7}$Division of Genetics/Dysmorphology, Rady Children's Hospital San Diego, San Diego, California, USA \\
		$^{8}$Division of Medical Genetics, A. I. du Pont Hospital for Children/Nemours, Wilmington, Delaware,USA
	}

	\IEEEauthorblockA { \\
Correspondence to: Yaron Gurovich, yaron@fdna.com.\\ 
 Boston 186 South St. 5th Floor, Boston, MA 02111 U.S.A., Tel: +1 (617) 412-7000	
}

	\IEEEauthorblockA { \\
	Conflict of interest: YG, YH, OB, NF, DG are employees of FDNA; LBS is an advisor of FDNA;  \\
	LBS, PK, LMB, KWG are members of the scientific advisory board of FDNA
}
}

\maketitle

\begin{abstract}
Facial analysis technologies have recently measured up to the capabilities of expert clinicians in syndrome identification.
To date, these technologies could only identify phenotypes of a few diseases, limiting their role in clinical settings where hundreds of diagnoses must be considered. 

We developed a facial analysis framework, DeepGestalt, using computer vision and deep learning algorithms, that quantifies similarities to hundreds of genetic syndromes based on unconstrained 2D images.
DeepGestalt is currently trained with over 26,000 patient cases from a rapidly growing phenotype-genotype database, consisting of tens of thousands of validated clinical cases, curated through a community-driven platform. 
DeepGestalt currently achieves 91\% top-10-accuracy in identifying over 215 different genetic syndromes and has outperformed clinical experts in three separate experiments. 

We suggest that this form of artificial intelligence is ready to support medical genetics in clinical and laboratory practices and will play a key role in the future of precision medicine.

\end{abstract}


\section{Introduction}

Diseases that are caused by underlying genetics affect a majority of people during their lifetimes. Here, we specifically assess syndromic genetic conditions. 
This class of diseases affects nearly 8\% of the population \cite{baird1988genetic}. 
Many affected individuals present symptoms that will affect their health and quality of life.
An early diagnosis is essential to prevent the occurrence of potential health problems, such as critical congenital heart diseases, respiratory problems, and developmental delays, among others.
It can also benefit the patients because special prevention and screening programs exist once the diagnosis has been established.

Many of these syndromes are known to have facial phenotypes \cite{hart2009genetic}, which are highly informative to clinical geneticists for diagnosing genetic diseases \cite{ferry2014diagnostically, basel2016recognition, rai2015using}. 
For the more common or distinctive syndromes, genetic experts are sometimes able to reach a diagnosis, or at least a strong hypothesis, based on the facial traits of the patient. 
However, in most cases, the patients see a genetic expert only years after the first symptoms occur. 
Often, due to the rarity of many syndromes and the large number of possible disorders, achieving the correct diagnosis involves a lengthy and expensive work-up that may take years or even decades \cite{kole2009voice} (the diagnostic odyssey). 

Recognition of non-classical presentations of common syndromes, or ultra-rare syndromes, may be constrained by the individual human expert's prior experience. 
The use of computerized systems as an aid or reference for clinicians is therefore becoming increasingly important. 
DeepGestalt holds the promise of making expert knowledge more accessible to healthcare professionals in other specialties such as pediatrics.

Recent advances in computer vision and machine learning, and specifically deep learning, present the opportunity for novel systems in many fields. 
In the last few years, the performance of tasks such as object detection, object localization, recognition and segmentation, on public datasets such as ImageNet \cite{ILSVRC15} and COCO \cite{lin2014microsoft} has dramatically improved. 
With this improvement, many new applications have emerged. 
As a result, the penetration of computer vision and artificial intelligence systems to commercial markets has increased. 
For example, this can be seen in autonomous driving projects, automatic detection of objects and faces on our mobile devices, advanced robotics and more. 

The facial phenotype is critical for syndrome diagnosis. 
To this end, facial analysis using computer vision has great potential. 
Computer vision research has long been dealing with facial analysis related problems. 
DeepFace \cite{taigman2014deepface} showed how Deep Convolutional Neural Networks (DCNNs) trained on large-scale data achieved human-level performance on the task of person verification on the Labeled Faces in the Wild (LFW) dataset \cite{LFWTech}. 
The work of \cite{parkhi2015deep} and \cite{yi2014learning-CASIA} boosted facial analysis research through the publishing of novel models and publicly available large-scale datasets.
Additional work, such as \cite{schroff2015facenet} and \cite{liu2015targeting} showed how facial DCNN models can be improved by using advanced loss functions and metric learning, setting new state-of-the-art results on the LFW dataset.
In \cite{levi2015age}, an automatic method for age and gender classification from facial images is proposed by using DCNNs. 
The work of \cite{liu2015deep} uses a cascade of two DCNNs in order to detect facial attributes, such as smiling, the presence of a mustache, eyeglasses etc. Another challenging task is recognizing facial expressions, such as surprise, sadness, happiness, fear etc. 
In \cite{ding2016facenet2expnet}, a method based on fine-tuning a DCNN is described and achieves good results using only a small dataset of samples.

The recognition of a genetic syndrome with a facial phenotype has many similarities to classic facial recognition.
However, in practice, developing a system for syndrome recognition is challenging for several reasons, such as limited data, subtle facial patterns and ethnic differences.
State-of-the-art face recognition systems are trained on large-scale datasets, starting from 0.5M images of thousands of people \cite{yi2014learning-CASIA}, and going up to 260M images of millions of people \cite{schroff2015facenet}.
The scale of these datasets is a key factor for the success of these systems.
It allows deep learning algorithms to learn robust and accurate models.
In the case of genetic syndromes, it is impossible to collect such a large dataset due to the rarity of these syndromes.
Potential datasets, are much smaller in scale, and are unbalanced, as is the large variance of patients per disease in the general population.
Another difficulty is the subtleness of the facial features in some syndromes, coupled with the fact that some syndromes do not show a characteristic facial phenotype that has been clinically described.

Most studies in the field of computer-aided syndrome recognition do not tackle the real world problem of classifying thousands of syndromes from unconstrained images.
They address problems like classifying unaffected from affected individuals, or diagnosing only one syndrome \cite{rai2015using}, usually using photos captured in a constrained manner.

In this report, we present a novel framework called DeepGestalt, which is one of the next-generation phenotyping technologies \cite{liehr2017next} used in the Face2Gene application (FDNA Inc., Boston, USA) to highlight phenotypes of thousands of diseases and millions of genetic variations. 
This technology significantly improves the process of recognizing genetic syndromes by enabling the robust recognition of hundreds of syndromes. 
It is proven to assist in the clinical setting (e.g. \cite{basel2016recognition,hadj2017automatic}), by offering meaningful insights for a large number of syndromes, based on the deep analysis of patients' facial images. 
There are also initial findings that the technology can be used to complement next generation sequencing (NGS) based molecular testing by inferring causative genetic variants from sequencing data \cite{gripp2016role}.

This report is organized as follows. Literature and related work review is given in Section \ref{s:Related Work}. 
Section \ref{s:Methods} reviews the technology methods, including the dataset usage, training paradigm and evaluation steps. 
In Section \ref{s:Experiments and Results}, we describe the experiments done for three types of syndrome classification tasks: Binary Gestalt Model, Specialized Gestalt Model and Multi-class Gestalt Model. 
We review the results achieved on these tasks and compare them, where possible, to other methods. 
Finally, in Section \ref{s:Conclusion} we discuss the results and summarize the ideas presented in this report.

\section{Literature Review}\label{s:Related Work}

We chose to survey four aspects of related work: A) Problem description B) Methods used C) Data used to train the system and D) Evaluation protocol in terms of data and results. 
A full comparison is given in Table \ref{tb:related_work}.

\subsection{Problem}
There are three main problems addressed in the literature: 
Problem 1: Single syndrome vs. other population - a binary classification problem of distinguishing subjects with a specific syndrome from normal (unaffected) subjects or subjects with other syndromes (\cite{shukla2017deep},\cite{saraydemir2012down},\cite{burccin2011down},\cite{zhao2014digital},\cite{kruszka2017down},\cite{basel2016recognition},\cite{kruszka201722q11},\cite{zhao2014ensemble},\cite{liehr2017next}). 
Problem 2: Syndromic vs. normal - a binary classification problem of distinguishing subjects with any syndrome from normal (unaffected) subjects (\cite{shukla2017deep},\cite{zhao2014digital},\cite{cerrolaza2016identification}). 
Problem 3: Multiple syndromes classification - a multi-class problem of identifying the correct syndrome from multiple possible syndromes (\cite{kuru2014biomedical},\cite{shukla2017deep},\cite{loos2003computer},\cite{boehringer2011automated},\cite{ferry2014diagnostically},\cite{boehringer2006syndrome}). Our work belongs to the third type, although we also demonstrate its capabilities on the first problem.

\begin{table*}[]
	\centering
	\caption{Related Work Summary}
	\label{tb:related_work}
	\begin{tabular}{|l|c|c|c|c|}
		\hline
		\multicolumn{1}{|c|}{}                                & \begin{tabular}[c]{@{}c@{}}Number of \\ Genetic Disorders\end{tabular}  & \begin{tabular}[c]{@{}c@{}}Number of Training \\ Samples (Syndromic)\end{tabular} & \begin{tabular}[c]{@{}c@{}}Evaluation Method\end{tabular}                                       & \begin{tabular}[c]{@{}c@{}}Accuracy \\ (top-1-accuracy)\end{tabular}            \\
		\hline 
		\multicolumn{5}{|c|}{\textbf{Problem 1: Single syndrome vs. other population}}    \\                                                                                      \hline                  
		Saraydemir et al. \cite{saraydemir2012down}         & 1    & 15         & 3,4-Fold Cross-Validation     & 97.34\%     \\ \hline
		Burccin et al. \cite{burccin2011down}               & 1    & 10         & 51 images in a test set        & 95.30\%      \\ \hline
		Zhao et al. \cite{zhao2014digital}                  & 1    & 50         & Leave-One-Out                  & 96.70\%      \\ \hline
		Basel-Vanagaite et al. \cite{basel2016recognition}            & 1    & 134        & 7 images in test set           & 94\%      \\ \hline
		Kruszka et al. \cite{kruszka2017down}               & 1    & 129         & Leave-One-Out                  & 94.30\%      \\ \hline
		Kruszka et al. \cite{kruszka201722q11}              & 1    & 156        & Leave-One-Out                  & 94.90\%      \\ \hline
		Liehr et al. \cite{liehr2017next}                   & 2    & 173        & 10-Fold Cross-Validation       & 100\%       \\ \hline
		Shukla et al. \cite{shukla2017deep}                 & 6    & 1126       & 5-Fold Cross-Validation       & 94.93 (mAP)\footnote[1]{}         \\ \hline
		Ferry et al. \cite{ferry2014diagnostically}         & 8    & 1363       & Leave-One-Out                  & 94.90\%      \\ \hline
		\multicolumn{5}{|c|}{\textbf{Problem 2: Syndromic vs. normal}}                                                                                                          \\ \hline
		Zhao et al. \cite{zhao2014digital}                  & 14    & 24         & Leave-One-Out                  & 97\%      \\ \hline
		Cerrolaza et al. \cite{cerrolaza2016identification} & 15   & 73         & Leave-One-Out                  & 95\%        \\ \hline
		Shukla et al. \cite{shukla2017deep}                 & 6    & 1126       & 5-Fold Cross-Validation       & 98.80\%      \\ \hline
		\multicolumn{5}{|c|}{\textbf{Problem 3: Multiple syndromes classification}}                                                                                                       \\ \hline
		Loos et al. \cite{loos2003computer}                 & 5    & 55         & Leave-One-Out                  & 76\%        \\ \hline
		Kuru et al. \cite{kuru2014biomedical}               & 15   & 92         & Leave-One-Out                  & 53\%        \\ \hline
		Boehringer et al. \cite{boehringer2006syndrome}     & 10   & 147        & 10-Fold Cross-Validation      & 75.70\%        \\ \hline
		Boehringer et al. \cite{boehringer2011automated}    & 14   & 202        & 91 images in a test set        & 21\%        \\ \hline
		Ferry et al. \cite{ferry2014diagnostically}         & 8    & 1363       & Leave-One-Out                  & 75.60\%\footnote[2]{}     \\ \hline
		Shukla et al. \cite{shukla2017deep}                 & 6    & 1126       & 5-Fold Cross-Validation       & 48\%\footnote[2]{}      \\ \hline
	\end{tabular}
\end{table*}

\subsection{Methods}
The most common methods consist of three stages: face and landmarks detection, feature extraction and classification.

\subsubsection{Face detection (FD) and facial landmarks detection}The purpose of this stage is to detect and localize the face within a given image (face detection) for alignment and localization purposes, and then detect specific facial landmarks, either for more accurate face alignment and localization, or for feature extraction around these landmarks. 
For FD, several works (\cite{ferry2014diagnostically, zhao2014digital, kruszka2017down, kruszka201722q11}) use the classical method of Viola-Jones \cite{viola2001rapid}. 
In \cite{ferry2014diagnostically}, Viola-Jones FD is followed by the method of Everingham \cite{everingham2009taking} for landmarks detection. 
The work of \cite{shukla2017deep} uses the method of Ramanan \cite{zhu2012face} for FD and the detection of 68 facial landmarks. 
In \cite{basel2016recognition}, a Haar-based cascaded face detector is used, followed by a 130 landmarks detector based on local image descriptors \cite{karlinsky2012using}.
\subsubsection{Feature extraction}In this part, we adopt the comparison approach of the survey paper by Rai et al. \cite{rai2015using}, which divides the feature extraction methods into holistic, local feature-based and statistical shape models, as explained below. 
We add DCNNs, which are used either for feature extraction or for direct classification.
\paragraph{Holistic methods}Holistic methods use a global representation based on the entire face. 
An example is Eigenface \cite{turk1991face}, which is based on Principal Component Analysis (PCA) as used by \cite{kuru2014biomedical}. 
Another example is Fisherface \cite{belhumeur1997eigenfaces}, which is based on Linear Discriminant Analysis (LDA). 
These methods are usually not competitive with state-of-the-art methods in face recognition tasks.
\paragraph{Local features based methods}Local features based methods represent the image by locally analyzing small image patches and aggregating the local information into a full image representation. 
Common methods are Scale-Invariant Feature Transform (SIFT) \cite{lowe2004distinctive} and Local Binary Patterns (LBP) \cite{ahonen2006face}, which is used in \cite{zhao2014digital, kruszka2017down, kruszka201722q11, burccin2011down, cerrolaza2016identification}. 
Other local appearance methods include Independent Component Analysis (ICA), Principal Component Analysis (PCA) and Gabor Wavelet Transform (GWT) and are used in \cite{zhao2014digital,ferry2014diagnostically, boehringer2006syndrome,loos2003computer, boehringer2011automated, saraydemir2012down}.
Another type of local features are geometric features, which analyze the geometric relations between facial landmarks, such as the normalized distance between landmarks, or the angles spanned by them. 
The works of \cite{zhao2014digital, kruszka2017down} and \cite{kruszka201722q11} use geometrical features, either in combination with texture features or without them.
\paragraph{Statistical model based methods}Statistical model based methods rely on the theory of statistical analysis of shapes. 
An example is the Active Appearance Model (AAM), which learns how the shape and texture of faces vary across the training images \cite{edwards1998face}. 
This method is used by \cite{ferry2014diagnostically} to detect eight different syndromes.
\paragraph{Deep Convolutional Neural Network methods}These methods are currently the most common for many computer vision tasks. 
Deep networks can be used for feature extraction, as done in \cite{shukla2017deep}, and can also be used for direct classification, as is presented in this work. 
DCNNs are very powerful classification models, when enough labeled data of the target domain has been used for training. 
In many cases, where such data do not exist, the networks are trained on an adjacent domain, and used for feature extraction or as a baseline for a knowledge transfer model to the target domain. 
The work of \cite{shukla2017deep} uses an object detection pre-trained Alex-Net model \cite{krizhevsky2012imagenet} as a baseline model. 
They then fine-tune the network on the LFW facial dataset and use the DCNN extracted features for the target domain of syndrome identification. In contrast, this study proposes a model that is designed to be trained directly on the target domain.

\footnotetext[1]{Mean Average Precision (mAP) is reported as in the original paper}
\footnotetext[2]{Calculated by us from the confusion matrix in the original paper} 

\subsubsection{Classification}
 Common methods include Support Vector Machine \cite{cortes1995support} as used in \cite{saraydemir2012down, burccin2011down, zhao2014digital, zhao2014ensemble, shukla2017deep}, K-Nearest Neighbors as used in \cite{saraydemir2012down, zhao2014ensemble, ferry2014diagnostically}, Deformable Models, Hidden Markov Models etc. Our model learns directly on the target domain and, therefore, we also use DCNNs as the classifier. 

\subsection{Training Data}Data quality and scale are key factors in computer vision and machine learning. 
The scale of the data becomes an acute problem for deep learning based methods. 
Table \ref{tb:related_work} reviews previous work with respect to the scale of the data used for training and also for evaluation.

\subsection{Evaluation}Since there is no public benchmark for comparison, it is impossible to compare the different methods in terms of accuracy.
In Table \ref{tb:related_work}, we describe the evaluation method used and the result reported for previous studies.
The table treats each problem separately, since a binary problem cannot be compared to a multi-class problem in terms of accuracy.
Most methods, other than \cite{ferry2014diagnostically, shukla2017deep}, are using small scale data (up to 200 train images).
The work of \cite{ferry2014diagnostically} and \cite{shukla2017deep} use over 1000 images for training, which still is considered to be a small number in the field of deep learning.
Although the performance can be evaluated by means of cross validation, the concern of over-fitting remains and the generalizability of the models is yet unclear.
In addition, the number of supported syndromes in a system is critical for clinical usage.
As shown in Table \ref{tb:related_work}, all previous works in the field, support a relatively small number of syndromes (15 or fewer).
In this work, we use a large dataset of tens of thousands of patient images, supporting hundreds of genetic syndromes, and evaluate it with a large external set collected from real clinical cases uploaded to Face2Gene from all over the world. 

\begin{figure*}[!h]
	\centering
	\includegraphics[scale=.41]{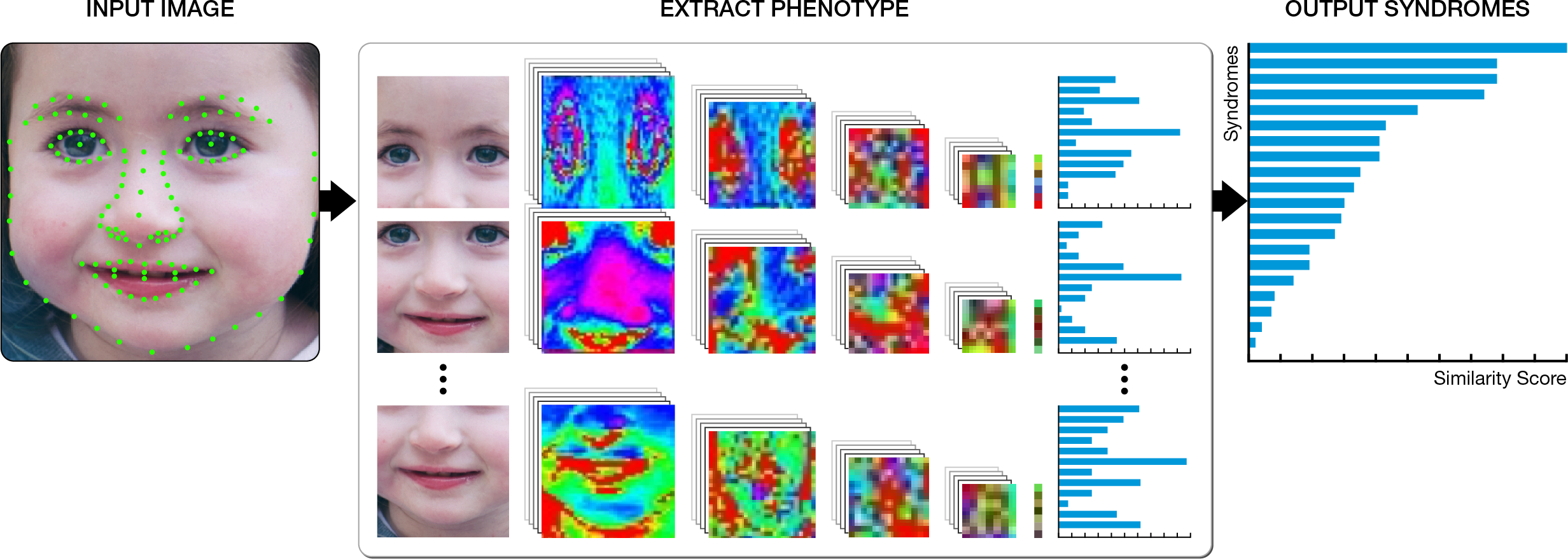}
	\caption{DeepGestalt: High level flow. The input image is first pre-processed to achieve a face detection, landmarks detection and alignment. After pre-processing, the input image is cropped into facial regions. Each region is fed into a DCNN to obtain a softmax vector indicating its correspondence to each syndrome in the model. The output vectors of all regional DCNNs are then aggregated and sorted to obtain the final ranked list of genetic syndromes. The histogram on the right hand side represents DeepGestalt output syndromes, sorted by the aggregated similarity score. Photo published with parental consent.}
	\label{fig:pipeline}
\end{figure*}

\section{Methods}\label{s:Methods}

This section describes the building blocks of the technology behind DeepGestalt.
We detail our image preprocessing pipeline, phenotype extraction and syndrome classification methods, datasets used, training details and evaluation protocol.
From an end-to-end perspective, our goal is to achieve a function $F(x)$, which maps an input image $x$ into a list of genetic syndromes with a similarity score per syndrome.
When sorted by this gestalt score, the top listed disorders represent the most likely differential diagnoses (Figure \ref{fig:pipeline}).

\subsection{Image Preprocessing}\label{ss:Image Preprocessing}

Our model is designed for real-world uncontrolled 2D images. 
The first step is to detect a patient's face in an input image. 
Since real clinical images have a large variance due to face size, pose, expression, background, occlusions and lighting, a robust face detector is needed in order to identify a valid frontal face.
We adopt a deep learning method, based on a DCNN cascade, proposed in \cite{li2015convolutional} for face detection in an uncontrolled environment. 
We adjust this method to fit our needs and operate optimally on images of children with genetic syndromes, in order to identify a frontal face from the image background.

We then detect 130 facial landmarks on the patient's face. 
This landmarks detection algorithm works in a chain of multiple steps, starting from a coarse step of identifying a small number of landmarks, up to a more subtle detection of all landmarks of interest \cite{karlinsky2012using}.

The resulting detected face and landmarks are first used to geometrically normalize the patient's face. 
The alignment of images reduces the pose variation among patients and shows improved performance on recognition tasks, such as face verification \cite{huang2012learning}.

The aligned image and its corresponding facial landmarks are then processed through a regions generator, which creates multiple pre-defined regions of interest from the patient's face. 
As illustrated in Figure \ref{fig:pipeline}, the different facial crops contain holistic face crops and several distinct regional crops which contain the main features of the human face, including the eyes, nose and mouth.
The final step in the preprocessing stage is to scale each facial cropped region to a fixed size of $100 \times 100$ and convert it to grayscale.

\subsection{Phenotype Extraction and Syndromes Classification}

In order to mitigate the main challenge of our specific problem, a small training database with unbalanced classes, we train the DeepGestalt model in two steps. 
First, we learn a general face representation and then fine-tune it into the genetic syndromes classification task.

\begin{figure*}[h!]
	\centering
	\includegraphics[scale=0.4]{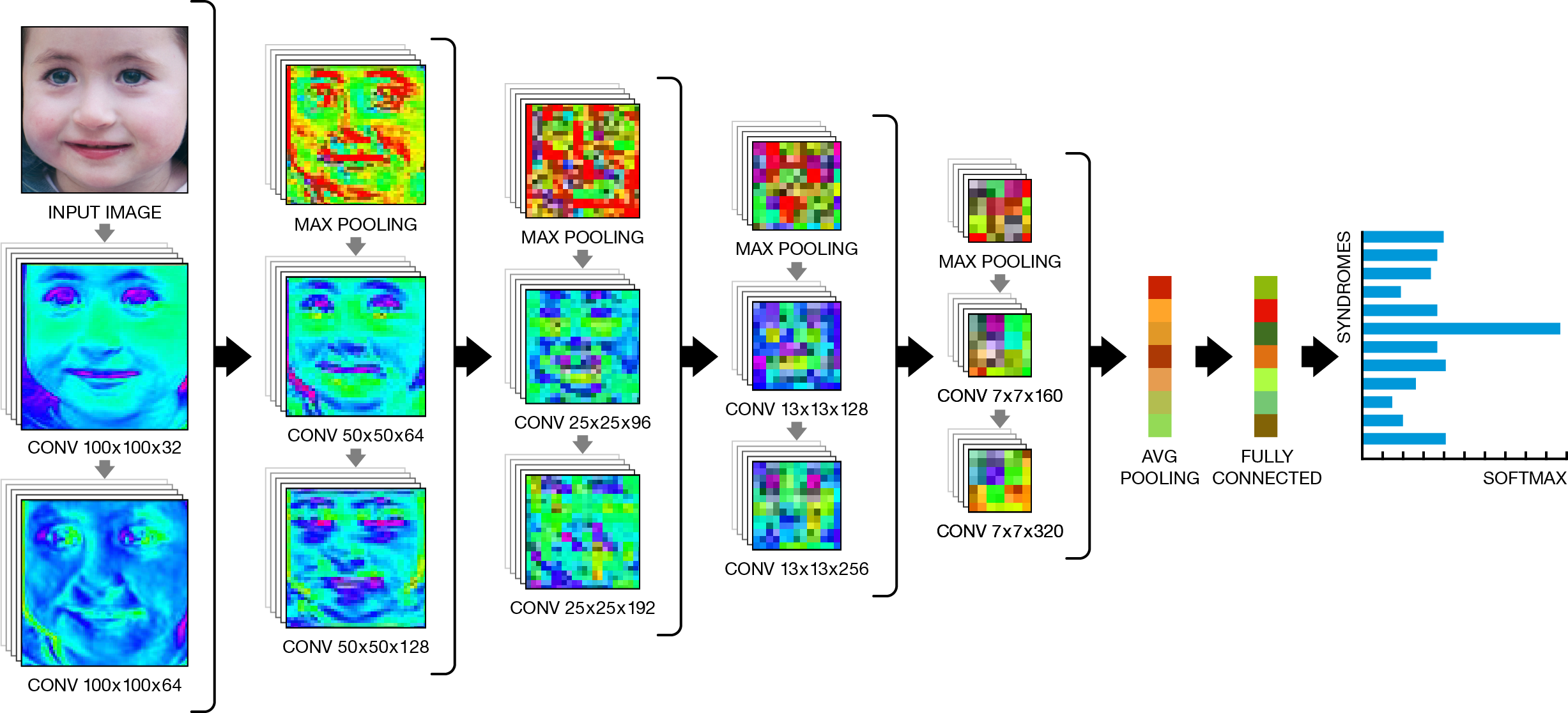}
	\caption{The Deep Convolutional Neural Network architecture of DeepGestalt. The network consists of ten convolutional layers and all but the last one are followed by batch normalization and ReLU. After each pair of convolutional layers, a pooling layer is applied (max pooling after the first four pairs and average pooling after the fifth pair). This is then followed by a fully connected layer with dropout (0.5) and a softmax layer. A sample heatmap is shown after each pooling layer. It is interesting to compare the low level features of the first layers with respect to the high level features of the final layers, where the latter identify more complex features in the input image and distinctive facial traits tend to emerge, while identity related features disappear. Photo published with parental consent.}
	\label{fig:net_spec}
\end{figure*}

To learn the baseline facial representation, we train a DCNN on a large-scale face identity database. 
Our backbone architecture is based on the one suggested in \cite{yi2014learning-CASIA} and is illustrated in Figure \ref{fig:net_spec}. 
We train separately for each facial crop, and combine the trained models to form a robust facial representation.

Once the general face representation model is obtained, we fine-tune the DCNN for each region with a smaller scale phenotype dataset for the task of syndrome classification. 
In practice, this step acts as a transfer learning step between a source domain (face recognition), and a target domain (genetic syndromes classification) \cite{yosinski2014transferable, taigman2015web}. 
Effectively, we use the powerful face recognition model for face representation (which performs comparably to the state-of-the-art results on the LFW benchmark \cite{LFWTech}), and train the model to separate different genetic syndromes, instead of separating different identities. 

We use the different facial regions, both as expert classifiers and as an ensemble of classifiers \cite{zhou2015naive,liu2015targeting}. 
Each region's specific DCNN separately makes a prediction, and these are combined by averaging the results and producing a powerful Gestalt model for a multi-class problem (Figure \ref{fig:pipeline}).

At the time of real clinical use, an image of a patient that has not been used during training is processed through the described pipeline. 
The output vector is a sorted vector of similarity scores, indicating the correlation of the patient's photo to each syndrome supported in the model.

\subsection{Datasets}

In order to train the model for face recognition, the publicly available CASIA Web-Face dataset \cite{yi2014learning-CASIA}, which contains 494,414 images from 10,575 different subjects, is aligned, scaled and cropped, as described above.

In order to fine-tune the networks to capture phenotypic information, we use a proprietary Face2Gene phenotype dataset - a validated dataset curated from Face2Gene users which includes tens of thousands of patient images with more than 2,500 genetic syndromes.

For system evaluation, we built a test set of real clinical cases.
We sampled, within a certain period of time, all real diagnosed clinical cases of any of the 216 syndromes supported at the time by DeepGestalt in the Face2Gene application. 
We automatically excluded images of low resolution and images where no frontal face was detected.
In addition, we removed images that were part of our training set and ignored duplicated images.
We ended up with 502 images covering 92 different syndromes.
The test set is skewed towards ultra-rare disorders, 66\% of the syndromes are represented in only 1 to 5 cases and 34\% in 6 to 39 cases. 
This results in a median value of 3.5 and average of 5.5 images per syndrome. This distribution of patients and syndromes mirrors the prevalence of rare disorders and is thus a representative test set for genetic counseling. 

Since Face2Gene keeps high standards of security and privacy, a fully automated processing system is used. 
Images are automatically processed within the same environment as they were uploaded by users, maintaining the privacy and security of those images. 
Only final results are reported, in order to evaluate performance.

\subsection{Training}
For each facial region, we train a face recognition DCNN using the large-scale face recognition dataset previously described. 
The dataset is randomly split into training (90\%) and validation (10\%). 
The region's networks are then fine-tuned for the genetic syndromes classification task using the Face2Gene phenotype dataset. 
This dataset is also split into training (90\%) and validation (10\%). 
The DCNNs architecture is inspired by the work of \cite{yi2014learning-CASIA}, with minor modifications, including adding batch normalization \cite{ioffe2015batch} layers after each convolutional layer (Figure \ref{fig:net_spec}). 

The training is done using Keras \cite{chollet2017keras} with TensorFlow as the backend \cite{tensorflow2015-whitepaper}. 
Baseline model training uses \textit{He Normal Initializer} \cite{he2015delving} weight initialization, which produced superior results compared to other known initializations. 
The optimization process uses \textit{Adam} \cite{kingma2014adam}, with an initial learning rate of $1e-3$ using a softmax loss function. 
After 40 epochs, we continue training the network for an additional 10 epochs using SGD with a learning rate of $1e-4$ and a momentum of 0.9. 

In the fine-tuning architecture, we replace the final layer output to match the number of syndromes in training. 
We found that the initialization for the fine-tuned layer is very important, and the best results are achieved when using a modified version of \textit{Xavier Normal Initializer} \cite{glorot2010understanding}. 
We experimented with different scales of \textit{Xavier Normal Initializer} and found that the best result was with a scale of 0.3. 

The fine-tuning optimizer is SGD with a learning rate of $5e-3$ and a momentum of 0.9. 
No weight decay or kernel regularization is used, since we found that the addition of batch normalization \cite{ioffe2015batch} to the original architecture \cite{yi2014learning-CASIA}, which also includes dropout (we set the rate to 50\%), performed better. 

Augmentation was proven to be significantly important. 
Each region is randomly augmented by rotation with a range of 5 degrees, small vertical and horizontal shifts (shift range of 0.05), shear transformation (shear range = $\frac{5\pi}{180}$), random zoom (zoom range = 0.05) and horizontal flip. 
Without augmentation, training quickly over-fitted, especially on the non-full-face regions.

Each region DCNN is independently trained for 50 epochs for the face recognition task and an additional 500 epochs for the fine-tune step.

\subsection{Evaluation}
We evaluate the model's performance by measuring the top-K-accuracy,  where $K=1, 5, 10$. 
For example, in the case of top-1-accuracy, the result represents whether the diagnosed syndrome was suggested first in the sorted list. 
The top-10-accuracy means that the correct genetic syndrome was suggested as one of the first ten syndromes in the sorted list.

In order to measure the statistical significance of our results for an unbalanced multi-class problem, we use a permutation test by measuring the distribution of the test set accuracy statistic under the null hypothesis. 
We randomly permute the test set labels $10^6$ times over the test data images, and calculate the top-K-accuracy for each of the permutations. 
This allows us to sample the accuracy distribution and to calculate its p-value.  

\subsection{Code availability}
DeepGestalt is the engine of the Face2Gene online application (www.face2gene.com), which is publically available to healthcare professionals.

\subsection{Data availability}
The data that support the findings of this study are divided into two groups, published data and restricted data.
Published data are available from the reported references.
Restricted data are curated from the Face2Gene application, and were used under license.
These images are not publicly available.

\section{Experiments and Results}\label{s:Experiments and Results}
 
\subsection{Binary Gestalt Model}

As described in Section \ref{s:Related Work}, many studies in the field of genetic syndrome classification deal with a binary problem, where the goal is to correctly classify unaffected individuals from affected ones, or to distinguish one specific syndrome from a mixed group of several other syndromes. 

In order to evaluate DeepGestalt on this type of binary problem, we train the model on only two cohorts. 
The first consists of patients' photos with a single syndrome (positive cohort) and the second consists of patients' photos with several different syndromes (negative cohort).

\subsubsection{Cornelia de Lange Syndrome (CdLS)}
We train the model using 614 CdLS images as the positive cohort, and 1079 images which are the negative cohort. 
The negative cohort images are of patients with several other syndromes (e.g. Kabuki, Aarskog, Dubowitz, Floating-Harbor, Fetal Alcohol, Kleefstra, and Rubinstein-Taybi syndromes). 

Following training, the model is evaluated on a test set that is described in \cite{basel2016recognition}.  
This test set includes 32 facial photos, 23 images of CdLS patients and 9 images of non-CdLS patients. DeepGestalt achieves 96.88\% accuracy in detecting the correct cohort.

We compare our result to previous studies conducted on those images (Table \ref{tb:cdls}). 
Basel-Vanagaite et al. \cite{basel2016recognition} reported an accuracy of 87\% in detecting whether or not the patient has CdLS. 
They also compared their method's performance to a previous study done by Rohatgi et al. \cite{rohatgi2010facial}, where those images were assessed by a group of 65 dysmorphologists achieving 75\% accuracy on the same task.

\begin{table}[h]
	\centering
	\caption{Results on the binary problem of detecting Cornelia de Lange syndrome patients}
	\label{tb:cdls}
	\begin{tabular}{|l|c|}
		\hline
		Method                                     & Accuracy \\ \hline
		Rohatgi et al. \cite{rohatgi2010facial}    & 75\%              \\ \hline
		Basel-Vanagaite et al. \cite{basel2016recognition} & 87\%              \\ \hline
		\textbf{DeepGestalt}               & \textbf{96.88\%}  \\ \hline
	\end{tabular}
\end{table}

\subsubsection{Angelman Syndrome (AS)}

This binary experiment focuses on separating Angelman Syndrome (AS) patients from patients with other syndromes (e.g. Williams, Russell-Silver, Fragile X, Moebius, DiGeorge, Mowat-Wilson, Aarskog, Chromosome 1p36 - Microdeletion, Prader-Willi, Kleefstra, Phelan-McDermid, Proteus, Feingold, Coffin-Siris). 
The model is trained using 766 AS images as the positive cohort, and 2699 images as the negative cohort. 

In a previous survey done by \cite{Bird201435}, a group of 20 dysmorphologists were asked to examine a set of 25 patient images and note which patients had AS and which did not.  
The test set included 10 patients with AS (positive cohort) and 15 patients with other genetic syndromes (negative cohort). 
However, experts were not aware of the number of patients in each cohort. 
The recognition rate reported in the survey was 71\% accuracy, 60\% sensitivity and 78\% specificity.

\begin{table}[h]
	\centering
	\caption{Results on the binary problem of detecting Angelman syndrome patients}
	\label{tb:as}
	\begin{tabular}{|l|c|c|c|}
		\hline
		Method                       & Accuracy      & Sensitivity   & Specificity    \\ \hline
		Bird et al. \cite{Bird201435}    & 71\%          & 60\%          & 78\%           \\ \hline
		\textbf{DeepGestalt} & \textbf{92\%} & \textbf{80\%} & \textbf{100\%} \\ \hline
	\end{tabular}
\end{table}

DeepGestalt was evaluated on the same test set and achieved a recognition rate of 92\%  accuracy, 80\% sensitivity and 100\% specificity (Table \ref{tb:as}), reducing the error rate by more than 72\%.

\subsection{Specialized Gestalt Model}

In this section, we describe how DeepGestalt may be used for a small scale problem, using only a small number of images per cohort. 
We focus on the problem of distinguishing between molecular subtypes of a syndrome which is genetically heterogeneous and derives from genetic errors in the same signaling pathway. 

We use this experiment as an example of a specialized Gestalt model, aimed at predicting the right genotype from patients with very subtle phenotype differences.

\begin{figure}[h]
	\centering
	\begin{tabular}{c@{\hskip 4pt}c@{\hskip 4pt}c@{\hskip 4pt}c@{\hskip 4pt}c}
		\small
		KRAS & PTPN11 & RAF1 & SOS1 & RIT1 \\
		\includegraphics[width=.18\linewidth]{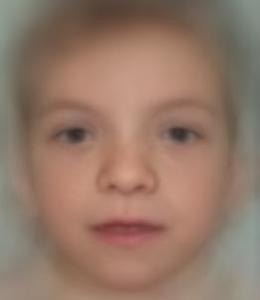}&
		\includegraphics[width=.18\linewidth]{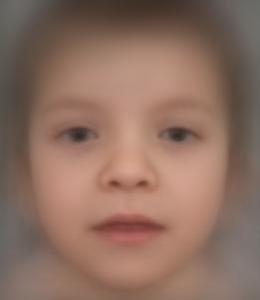}&
		\includegraphics[width=.18\linewidth]{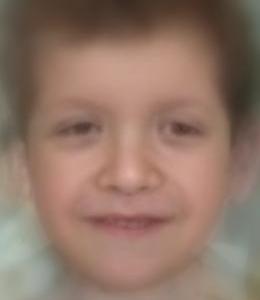}&
		\includegraphics[width=.18\linewidth]{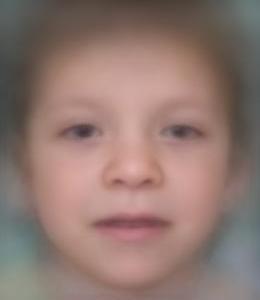}&		
		\includegraphics[width=.18\linewidth]{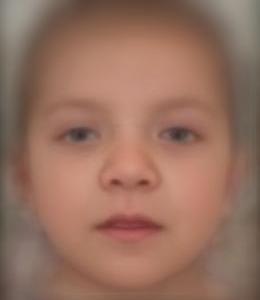}\\		
		
	\end{tabular}
	\caption{Composite photos of Noonan syndrome patients with different genotypes show subtle differences, such as less prominent eye brows in individuals with a SOS1 mutation, which might reflect the previously recognized sparse eye brows as an expression of the more notable ectodermal findings associated with mutations in this gene.}	
	\label{fig:Masks_Images_Correction_Final}
\end{figure}

In 2010, Allanson et al. published \textit{The face of Noonan syndrome: Does phenotype predict genotype} \cite{allanson2010face}. 
They explored whether dysmorphology experts can predict the Noonan syndrome related genotype using the facial phenotype. 
They presented a set of 81 images of Noonan syndrome patients to two dysmorphologists. 
The patients' genotypes have been confirmed molecularly as \textit{PTPN11, SOS1, RAF1} and \textit{KRAS}. 
The task was to predict the right genotype from a facial image. 
Their conclusion was that experts in the field could not succeed in this task, as written in the article abstract: ''Thus, the facial phenotype, alone, is insufficient to predict the genotype, but certain facial features may facilitate an educated guess in some cases''. 

We aim to examine if the technology described in this paper can perform better and propose a novel way to harness the Gestalt model technology to solve the problem of predicting the right genotype. 

\begin{figure}[h]
	\centering
	\includegraphics[scale=0.48]{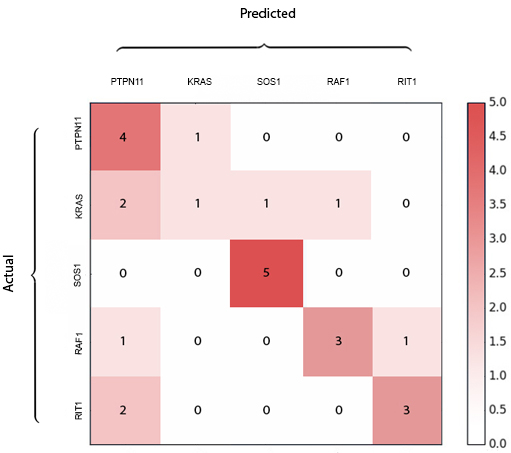}
	\caption{Test set confusion matrix for the Specialized Gestalt Model}
	\label{fig:ns_conf_mat}
\end{figure}

We collected patient images diagnosed with Noonan syndrome and molecularly diagnosed with a mutation in one of the following genes: \textit{PTPN11, SOS1, RAF1, RIT1} and \textit{KRAS}.
All of the images were annotated by experts, taken either from published articles or the internal Face2Gene phenotype database. 
A set of 25 images, five images per gene (type), are excluded from training and used as a test set. 
Those images were curated from \cite{gulec2015novel,zenker2007sos1,allanson2010face,rusu2014genotype,cave2016mutations,kouz2016genotype}. 
To illustrate the general appearance of each cohort, we create composite photos by averaging the images of each cohort, see Figure \ref{fig:Masks_Images_Correction_Final}.

Using the framework described above, we use this specialized dataset along with our internal dataset and train a full DeepGestalt Model.
The specialized Gestalt model is a truncated version of the full model and predicts only the five desired classes.
The resultant model is then applied to the test set and achieves a top-1-accuracy of 64\%. 
This is more than three times better than the random chance of 20\%. 
The confusion matrix for this test set is presented in Figure \ref{fig:ns_conf_mat}. 
A similar work using our technology with comparable results can be seen in \cite{zenkerESHG2017}.

Besides the phenotypes that are caused by mutations in the MAPkinase pathway, DeepGestalt has also been used to analyze two further molecular pathway diseases that are known for their high phenotypic similarity. 
In GPI-anchor biosynthesis deficiencies, DeepGestalt was able to reproduce the phenotypic substructure that was already delineated by expert clinicians and, beyond that, to deduce significant gene-specific phenotypes \cite{Knaus216291}. 
For five metabolic disorders of high similarity, amongst them Mucopolysaccharidosis type I and II, DeepGestalt was able to achieve accuracies in the differentiation of patient photos that are significanlty better than randomly expected \cite{Pantel219394}. 

Specialized Gestalt Modelling by DeepGestalt therefore has the potential to assist syndromoligists in the delineation of characteristic facial features in Mendelian diseases. 
\subsection{Multi-class Gestalt Model}

DeepGestalt is designed to perform facial Gestalt analysis at scale - supporting hundreds or potentially even thousands of different syndromes, as required in the clinical setting. 
It is trained on the validated Face2Gene phenotype dataset, which is rapidly growing by the large community of Face2Gene users.
To utilize the growth of the database, DeepGestalt is trained periodically and is evaluated on real clinical cases.

The results presented in this section are achieved by a model trained on a snapshot of the dataset, supporting 216 different syndromes and using 26,190 images derived from the full set of images in the current database. 
The model is evaluated on a test set of 502 images of real clinical cases.

This experiment includes the largest training dataset, largest amount of syndromes and largest test set, compared to all other methods published in this field (see Table \ref{tb:related_work}). 

DeepGestalt uses an aggregation of facial regions in order to achieve better performance and to improve robustness. 
This aggregation is forcing a majority vote paradigm over the different facial regions. 
To examine how each region contributes to the final model, we run a test set evaluation on each region separately and compare it to the aggregated model. 
As shown in Table \ref{tb:results}, the aggregated model performs better than each of its components (for simplicity purposes, we choose to present only top-5 accuracy in this table, top-1 and top-10 accuracy show the same trend). 
This emphasizes the ability of the proposed model to learn complementary phenotype information on different facial regions. 

\begin{table}[h]
	\centering
	\caption{Performance comparison between different facial regions and the final aggregated DeepGestalt model. The aggregated accuracy is a result of all regions as an ensemble of predictors.}
	\label{tb:results}
	\begin{tabular}{lcc} \toprule
		Facial Area                & \multicolumn{1}{l}{Top-5-Accuracy}  \\ \midrule
		Eyes                              & 60.36                                          \\
		Nose                              & 64.94                                                       \\
		Middle face (Ear to Ear)          & 69.92                                    \\
		Upper Half Face                   & 71.12                                        \\
		Lower Half Face                   & 62.15                                          \\
		Full Face                         & 77.49                                                \\
		&                                &                                    \\
		{\textbf{Aggregated model}}   & {\textbf{83.70}}             \\\bottomrule		
		\hline                        
	\end{tabular}
\end{table}

A full evaluation of the proposed model is shown in Table \ref{tb:perm_test} with the permutation test results. 
DeepGestalt achieves top-K-accuracy of 60.0\%, 83.7\% and 91.0\% for $K=1, 5, 10$ respectively on the test set. 
Permutation test mean accuracy and standard deviation ($\mu \pm \sigma$) are $ 3.0\% \pm 0.73 $, $ 10.4\% \pm 1.27 $ and $ 17.8\% \pm 1.55$ for $K=1, 5$ and $10$ respectively.
This yields a p-value equal to zero for all top-K-accuracies. 
This result shows a high level of statistical significance and strengthens the confidence that this model performs very well on such a challenging evaluation dataset of real clinical cases.

\begin{table}[h]
	\centering
	\caption{DeepGestalt Performance and permutation test results}
	\label{tb:perm_test}
	\begin{tabular}{|c|c|c|c|}
		\hline
		& \begin{tabular}[c]{@{}c@{}}Model\\ Accuracy\end{tabular} & \begin{tabular}[c]{@{}c@{}}Permutation Test \\ Mean Value\end{tabular} & \begin{tabular}[c]{@{}c@{}}Permutation Test\\ SD Value\end{tabular} \\ \hline
		top-1-accuracy  & 60.0\%                                                     & 3.0\%                                                                  & 0.73                                                                \\ \hline
		top-5-accuracy  & 83.7\%                                                   & 10.4\%                                                                 & 1.27                                                                \\ \hline
		top-10-accuracy & 91.0\%                                                     & 17.8\%                                                                & 1.55                                                                 \\ \hline
	\end{tabular}
\end{table}

\section{Discussion}\label{s:Conclusion}

This work presents a novel facial analysis framework for genetic syndrome classification called DeepGestalt. 
The proposed framework leverages deep learning technology and learns facial representation from a large-scale face recognition dataset. 
It is followed by a fine-tune phase, in which this knowledge is transferred to the genetic syndrome domain.

We show how this framework is able to generalize well for specific problems. 
We demonstrate how a binary model, trained to identify a single syndrome, surpasses human expert performance, both on Cornelia de Lange syndrome patients and on Angelman syndrome patients.

We present a specialized gestalt model, which focuses on the problem of identifying the correct facial phenotype on five genes related to the Noonan syndrome. 
This demonstrates the ability to generalize from small datasets and enables genetic experts to investigate new phenotype-genotype correlations and to gain novel clinical insights. 

Finally, we show DeepGestalt's high discrimination performance on hundreds of genetic syndromes characterized by unbalanced classes distribution. 
The evaluation on a large set of patients' photos achieves 91\% accuracy including the correct diagnosed syndrome as a top-10 list out of hundreds of possible genetic syndromes. 
This evaluation was done on 502 images of randomly sampled cases uploaded to Face2Gene, and follows our intention to simulate a close to clinic evaluation scenario. 
We, therefore, believe that this accuracy represents a common clinical use case, taking into account the prevalence and diagnostic challenges of different syndromes.

The increased ability to describe phenotype in a standardized manner opens the door to the emerging field of precision medicine, as well as to the  identification of new genetic syndromes by matching undiagnosed patients sharing a similar phenotype. 
Patient matching with such Artificial Intelligence (AI) technology will enhance the way that genetic syndromes and other genetically caused diseases are studied and explored.

In future studies, we aim to combine the Gestalt model described here with genome sequencing data. 
This will enable improved prioritization of gene variant results. 
It is the authors' belief that the coupling of the phenotype analysis, done by computer vision algorithms, with the continuously growing genomic knowledge, will open new ways to rapidly reach an accurate molecular diagnosis for patients with genetic syndromes, and may become a key-factor for the field of precision medicine.

\hfill

\hfill

\section*{Acknowledgment}
The authors would like to thank the patients and their families, as well as Face2Gene users worldwide who contribute with their knowledge and dedication to the improvement of this and other tools for the ultimate benefit of better health care.

\bibliographystyle{ieeetr}
\bibliography{references}

\end{document}